# AI visualization in Nanoscale Microscopy

A. Rajagopal[1] [0000-0002-5602-6416], V. Nirmala[2][0000-0001-7370-2490], Andrew.J[3] [0000-0003-3592-6543], Arun Muthuraj Vedamanickam[4][0000-0001-5361-9438]

[1] Indian Institute of Technology, Madras, India
[2] Queen Marys College, Chennai, India (corresponance)
[3] Karunya Institute of Technology and Sciences, India
[4]National Institute of Technology, Trichy, India
[1]rajagopal.motivate@gmail.com, [2]gvan.nirmala@gmail.com, [3]andrewj@karunya.edu, [4]arun.gvan@gmail.com

**Abstract.** Artificial Intelligence (AI) & Nanotechnology are promising areas for the future of humanity. While Deep Learning based Computer Vision has found applications in many fields from medicine to automotive, its application in nanotechnology can open doors for new scientific discoveries. Can we apply AI to explore objects that our eyes can't see such as nano scale sized objects? An AI platform to visualize nanoscale patterns learnt by a Deep Learning neural network can open new frontiers for nanotechnology. The objective of this paper is to develop a Deep Learning based visualization system on images of nanomaterials obtained by scanning electron microscope (SEM). This paper contributes an AI platform to enable any nanoscience researcher to use AI in visual exploration of nanoscale morphologies of nanomaterials. This AI is developed by a technique of visualizing intermediate activations of a Convolutional AutoEncoder (CAE). In this method, a nano scale specimen image is transformed into its feature representations by a Convolution Neural Network (CNN). The Convolutional AutoEncoder is trained on 100% SEM dataset from NFFA-EUROPE, and then CNN visualization is applied. This AI generates various conceptual feature representations of the nanomaterial.
While Deep Learning based image classification of SEM images are widely published in literature, there are not much publications that have visualized Deep neural networks of nanomaterials. There is a significant opportunity to gain insights from the learnings extracted by machine learning. This paper unlocks the potential of applying Deep Learning based Visualization on electron microscopy to offer AI extracted features and architectural patterns of various nanomaterials. This is a contribution in Explainable AI in nano scale objects, and to learn from otherwise black box neural networks. This paper contributes an open source AI with reproducible results at URL (https://sites.google.com/view/aifornanotechnology)

**Keywords:** Explainable AI, Deep Learning in Microscopy, Convolutional AutoEncoder, CNN visualization, Nanomaterials

# 1 Introduction

## 1.1 The opportunity: Applications of AI to new fields

Artificial Intelligence (AI) & Nanotechnology are transforming science & technology. The potential to apply AI is immense and rapidly progressing across many fields, but this potential is not widely used today by nanoscience. An editorial in highly reputed nature methods journal highlighted the potential of Deep Learning in Microscopy. Deep Learning Computer Vision approaches have established broad set of applications in many industries such as healthcare to Self-driving cars. There is an untapped opportunity for application of AI to nano-scale objects. To unlock this potential, this paper explores AI visualization techniques in nano scale objects.

## 1.2 Research gap

Deep Learning in microscopy has a significant potential as per the editorial of Nature Methods (Editorial, "Deep learning gets scope time", 2019). As per methods to watch in Nature Methods (Strack, 2018), an astonishing use of Deep Learning is not image analysis, but "image transformation" (Moen et al., 2019). As per Mar 2021 topic review of Deep Learning in microscopy (Ede, 2021), the current literature is limited to tasks such as image classification, image segmentation, image reconstruction. But tasks such as image visualization of CNN feature maps of nanomaterials is not well published. This paper contributes into this gap. To the best of our knowledge, this paper is the first in literature to develop an AI visualization toolkit on nanomaterials. Specifically, this is the first work to apply Deep Learning visualization techniques presented by (Zeiler & Fergus, 2014) and (Chollet, 2018) on nanoscale materials. While CNN feature visualization proposed by Zeiler & Fergus (2014) is widely popular in Machine Learning literature, its applications on nanomaterial dataset is not well published so far.

## 1.3 Contributions

The significance of the contribution is towards democratizing access to Deep learning for the benefit of the research community by contributing an Open Access AI based visualization toolkit, NanoAID. This paper contributes this AI in open source software. All results are reproducible at https://sites.google.com/view/aifornanotechnology. As summarized in Table 1, the contributions are

1. The Open Source AI platform for enabling nanoscience community, NanoAID.
2. The AI equips nano scientists to study nanostructures as CNN filters transform a SEM image of a nanomaterial to its feature representation (Example in Fig 1). An intuitive concept of "AI lens" is proposed in Fig 1 & demonstrated in Fig 3.
3. In addition, this AI equips the community to visualize the CNN extracted architectural characteristics for a class of nanomaterial. (Example in Fig 2).
4. Future extensions of NanoAID could include domain applications as shown in Fig 5 and Fig 6.

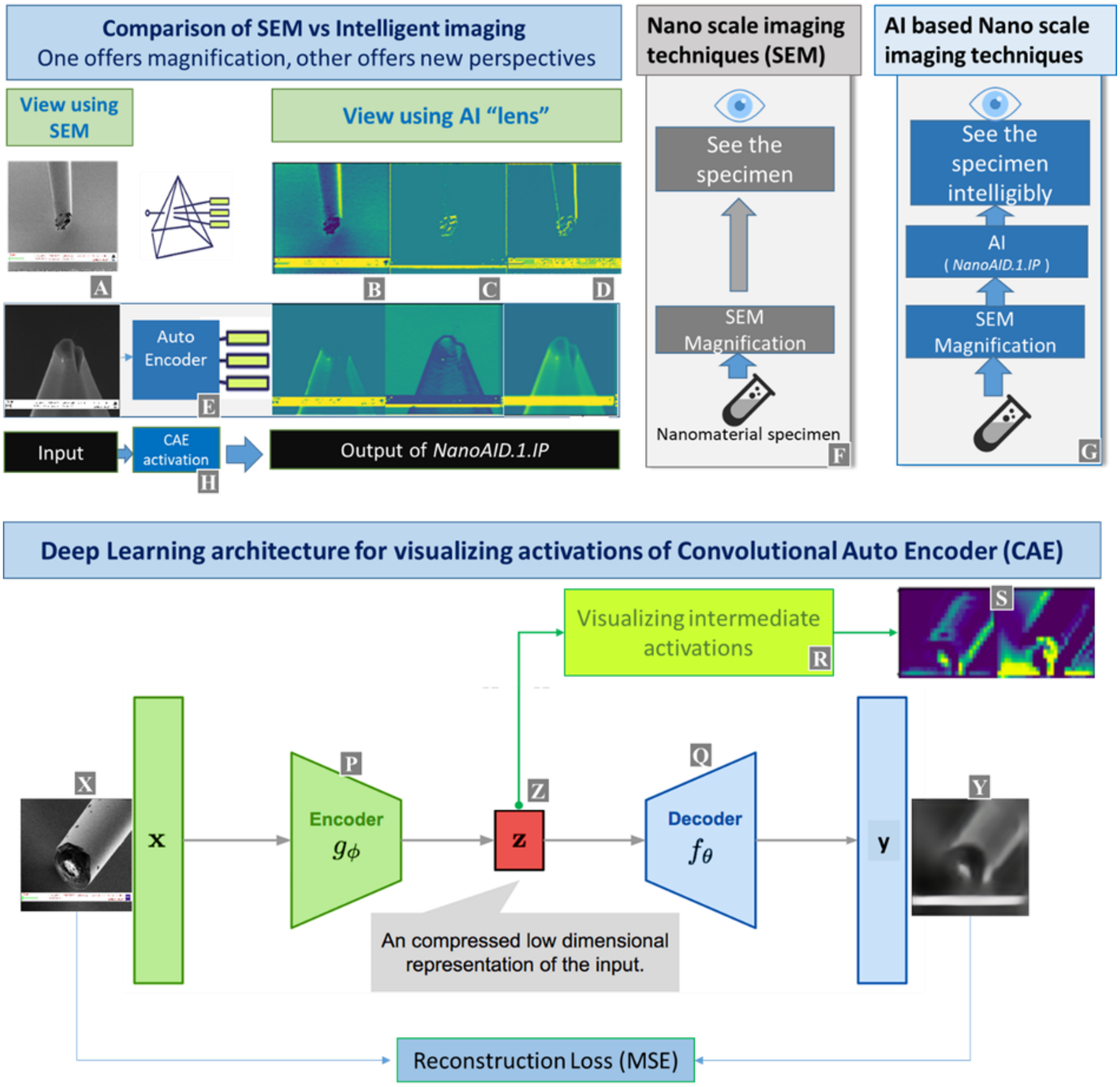

**Figure 1:** *NanoAID.#1.IP* module offers a AI based nanoscale imaging method. This method uses gradient ascent in input space to visualize intermediate activations of filters of a Convolutional AutoEncoder. The intuition of proposed "AI lens" based idea for intelligent imaging is illustrated in (G) in comparison with classical Scanning Electron Microscopy (F). Using visualizing intermediate activations of a Convolutional AutoEncoder (P)-(Q)-(R), it is possible to re-represent a nano scale specimen such as (A) into feature representations such as (B), (C), (D).

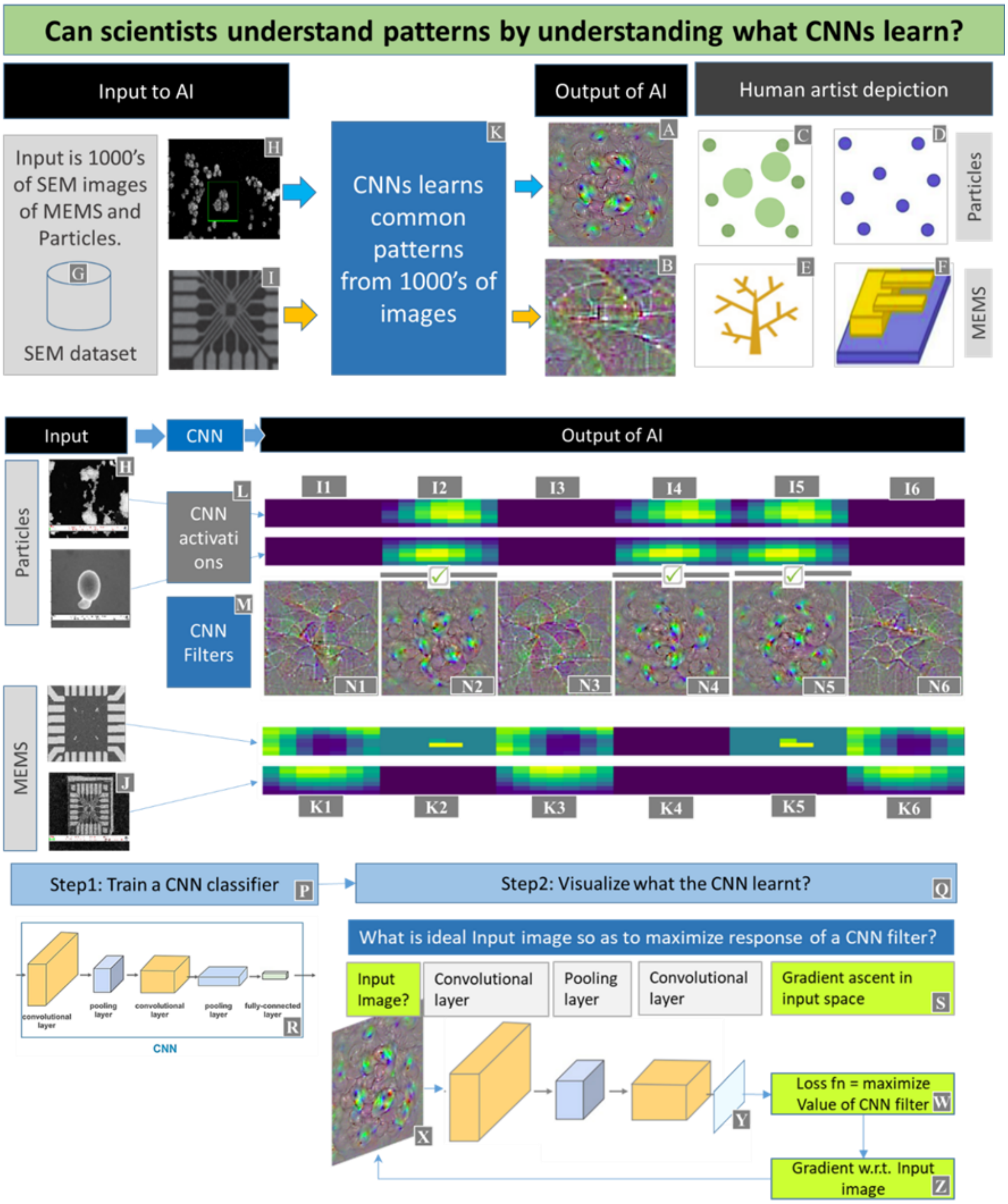

**Figure 2:** *NanoAID.#2* module explores the potential of AI to visualize patterns that a neural network has learnt that is common to thousands of SEM images of a particular class of nanomaterial. By visualizing the patterns a convolutional filter would respond to maximally, this AI demonstrates it is possible to gain understanding of the common

features of a class of nanomaterials. The top section of picture shows the general idea of using CNNs to learn common patterns. Next the middle section shows the output of NanoAID. The bottom section of the picture describes the neural network architecture used to generate the output. This output was generated by performing these two steps. In Step 1, a CNN classifier (R) is trained on SEM dataset as shown in (P). In Step 2, convolutional filters are visualized as shown in (Q).

### 1.4 Literature review & Novelty

There is amazing progress in Deep Learning in microscopy of biological images obtained with electron microscope. As per recent article in Nature Communications, researchers have developed AI toolkit for nanoscale bio images (Von Chamier et al., 2021). Neural network based Morphological analysis of nanometer objects are less established as per Nature Communications article (Schubert et al., 2019), yet this paper contributes in this aspect. While user friendly AI tools are developed & reported in reputed Nature journals by (Berg et al., 2019) and (von Chamier et al., 2021), these are on nanoscale images of biological organisms. This paper also contributes to AI tool, but the special thing about this tool is CNN visualization on many nanomaterial classes such MEMS devices, Tips, Patterned surfaces, Particles, Fibers.

A Convolution Neural Network (CNN) based image classification task is reported in Nature Scientific Reports (Modarres et al., 2017). While Modarres et al. (2017) develop an image classification by "black box" CNN classifier, this paper utilizes the same dataset, but performs a different task of CNN based visualization to understand what the Deep Learning "black box" is learning. As per 2019 (Editorial, "Deep learning gets scope time", 2019), understanding the deep learning "black box" is an active area of research. The dataset utilized in this paper is 100% SEM dataset, a publicly available dataset of Scanning Electron Microscopy (SEM), published in Nature Scientific Data by NFFA–EUROPE project (Aversa et al., 2018).

The contribution is in paper assumes significance in the context of Explainable AI for advancing nanotechnology. The recent paper by (Borowski et al., 2021) at International Conference on Learning Representations (ICLR 2021) discuss why CNN feature visualizations are valuable technique for Explainable AI. Many researchers support the view that the feature maps such as one presented in Figure 2 are meaningful (Olah et al., 2020). The "AI microscope" developed by leading AI research lab at https://microscope.openai.com/models, OpenAI (Schubert 2020) demonstrate the potential of Explainable AI, and this paper extends this idea to nanomaterials.

While electron microscopy has advanced nanotechnology research, the capability to view nanomaterials through an "AI lens" (illustrated in Figure 1) is pioneered by this research paper. While Scanning Electron Microscopy (SEM) opened the doors for us to see previously unseen structures, this paper proposes an "AI lens", that opens the doors for the community to see previously unseen abstract structural features of nanomaterials.

# 2 Methods & Results

| Reproducible Results | | URLs |
|---|---|---|
| **1)** Contributed an AI platform for benefit of the community | NanoAID opens doors the opportunity to apply CNN visualization on nano-material. | NanoAID website |
| **2)** Demonstrated novel imaging technique in SEM by applying AI | "AI lens": Nanomaterials are seen through CNN filters to visualize features. (Fig 1, Fig 3) | URL#1 (click) |
| **3)** Explored patterns learnt by machine | Visualize common patterns of nano-materials by learning patterns from thousands of SEM images (Fig 2, Fig 4) | URL#2 |
| **4)** Future applications | Patterns discovered by AI in MEMS and Patterned Surfaces (Fig 5, Fig 6) | URL#3<br>URL#4 |

*Table 1 : Summary of results*

## 2.1 Open Source AI framework to enable discoveries in Nanotechnology

Deep Learning has been established in representing objects at human scale such as the common day objects found in IMAGENET dataset. Is it possible to extend this success of Deep Learning methods to explore nano-scale objects?

Nanotechnology deals with objects with nanoscale structures with a length scale usually cited in nanometers. One nanometer is one-billionth of a meter. A sheet of paper is about 100,000 nanometers thick! The goal of this paper is to enable every scientist to apply AI visualization to advance the nanoscience domain and help succeed in future applications of AI in nano. With this design goal, we created Nano scientist's AI Discovery services, NanoAID, and bestowing this effort as an Open Source AI platform. All the results are reproducible online, and shared in the form of Google colab URLs for ease of reproducing results.

## 2.2 Seeing a nanomaterial through a "AI lens"

***Result:*** This technique enables a researcher to view any given SEM image through the eyes of a neural network, specifically through the feature maps of an AutoEncoder. The Autoencoder neural network will learn the most essential features of a nanomaterial given the training objective is to minimize the reconstruction loss function, thus by visualizing the intermediate activations of a Convolutional AutoEncoder, it is possible to highlight the essential features of an nanomaterial specimen. As illustrated in Fig No 1, this method is able to transform any SEM image (A) into feature maps shown in (B), (C), (D). This transformed image is now visualised (R) and presented to the user. Essentially the Encoder neural network performed information distillation, and thus re-represented the input image (A) into newer perspectives (B), (C), (D). Given AI based visualization is the primary output, key results of this paper are in visual format, hence the figures are best viewed in digital format.

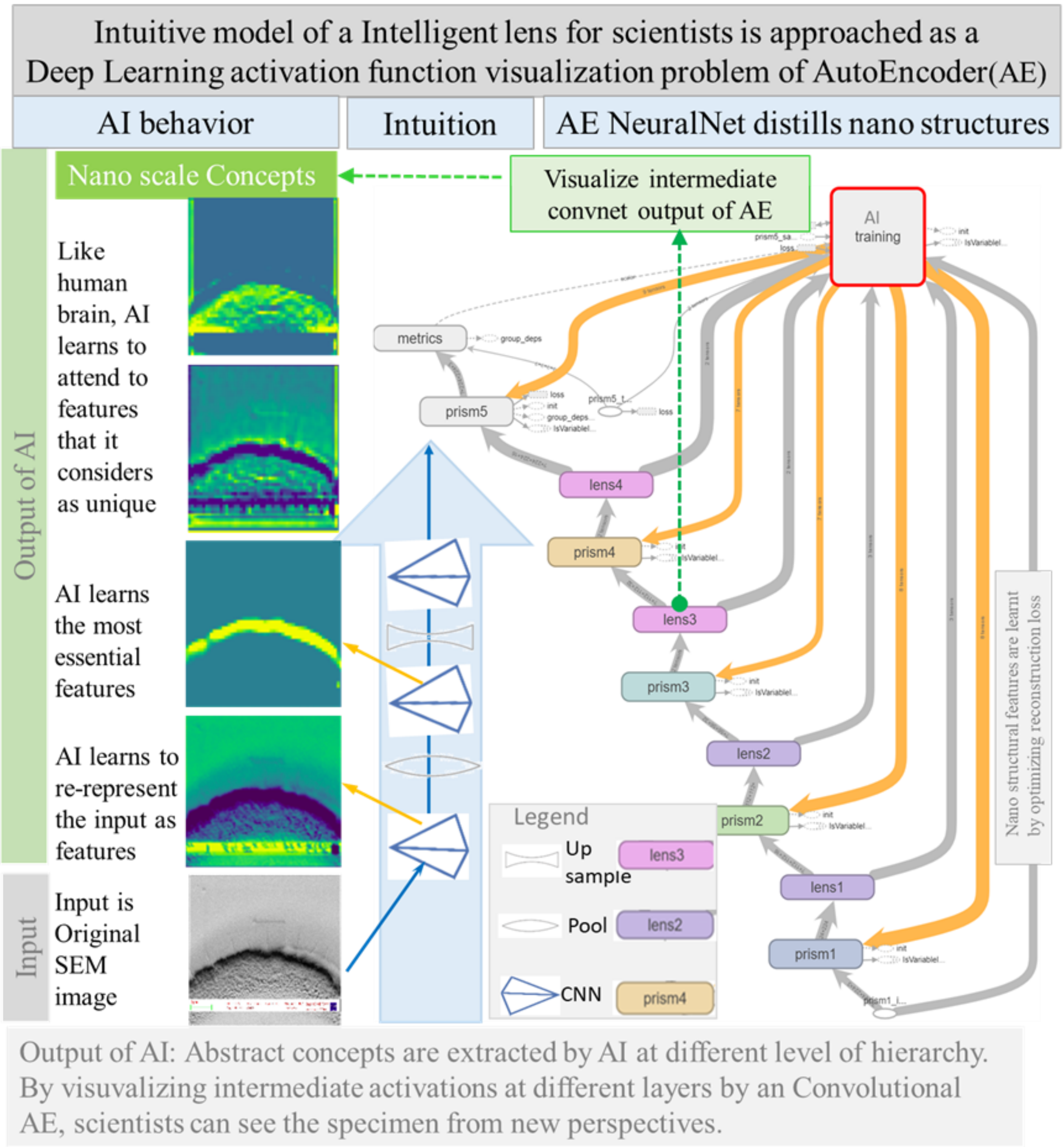

**Figure 3:** The intuition of "AI lens" is illustrated. A Tensorflow computation graph consisting of CNN layers and pooling/ upsampling layers of a CAE is presented on the right side of this figure. By visualizing output of intermediate Convnet layers, the input SEM image is transformed into its feature representations, shown in green color images in the left.

***Method:*** The method is illustrated in Fig No 1 in the neural network architecture. In this method, first a Convolutional AutoEncoder is trained on 100% SEM dataset (Aversa et al., 2018) to minimize the pixel wise Mean Square Error (MSE) reconstruction loss between the input image(X) and the reconstructed image (Y) in Fig No 1. An appropriate bottleneck size for the latent vector (Z) was choosen to ensure the reconstruction loss was at reasonable level by observing the reconstructed image. Then the Encoder part(P) of the Autoencoder is picked up and saved as a neural network model. The User Interface (UI) in the URL allows the researcher to select the number of layers in the Encoder to transform an input image. Based on the selection in the UI, a neural network model is dynamically created with the selected depth. The neural network model is created on the fly using the TensorFlow/Keras Functional API (Chollet, 2018). Using this dynamically created model, the input SEM image is transformed by the various activation functions of this dynamically created model. This CNN visualization method was introduced by Zeiler & Fergus (2014) and implemented in Keras by Chollet (2018). Researchers are beginning to apply this CNN visualization method in other fields such as MRI as per nature scientific reports (Oh Chung at al., 2019).

### 2.3 Visualizing nano scale patterns learnt by CNN

***Result:*** As articulated in Table 1, the second result is an AI to see commonly occurring patterns in a class of nanomaterials. This AI module allows nanotechnologists to extract common structural of MEMS and nanoparticles, by coherent analysis of thousands of SEM images. This is demonstrated in Fig 2. This problem was approached by learning the common features into a Convolutional Neural Network by training to learn the common patterns, and then visualizing its convent layers to visualize the learnings.
This AI equips nano scientists to comprehend structural patterns that are typically to a class of nanomaterial. This helps in explainable classification in contrast to black box approach used in the nature journal (Modarres et al. 2017). The NanoAID website https://sites.google.com/view/aifornanotechnology/ displays the patterns learnt by AI. This has the potential to open up the science of understanding of morphology of various nano scale objects in nanotechnology. As depicted in Fig 2, (A) and (B) shows the patterns of NanoParticles and MEMS respectively. These patterns were learnt by the CNN classifier (R) as the CNN learns from thousands of representative SEM images of NanoParticles and MEMS from the 100% SEM Dataset (Aversa et al., 2018). With a close examination of Fig 2, it can be noticed that the activations for MEMS happen in (K1, K3, K6) due to CNN filters (N1, N3, N6), while for Nanoparticles are activated by CNN filters (N2, N4, N5). This deep insight can be valuable for future applications that base upon Explainable AI for nanoscience.
***Method:*** The paper also explored three different methods to train the neural network & visualize the learnings. As seen in Fig 4, the patterns were visualized by experimenting 3 different training methods: classical transfer learning based training (A1), transfer learning with fine tuning all layers (A2), train a randomly initiated network (A3). The work by researchers at OpenAI demonstrate patterns learnt from IMAGENET at https://microscope.openai.com/models, and this paper tailor develops this concept for NanoMaterials.

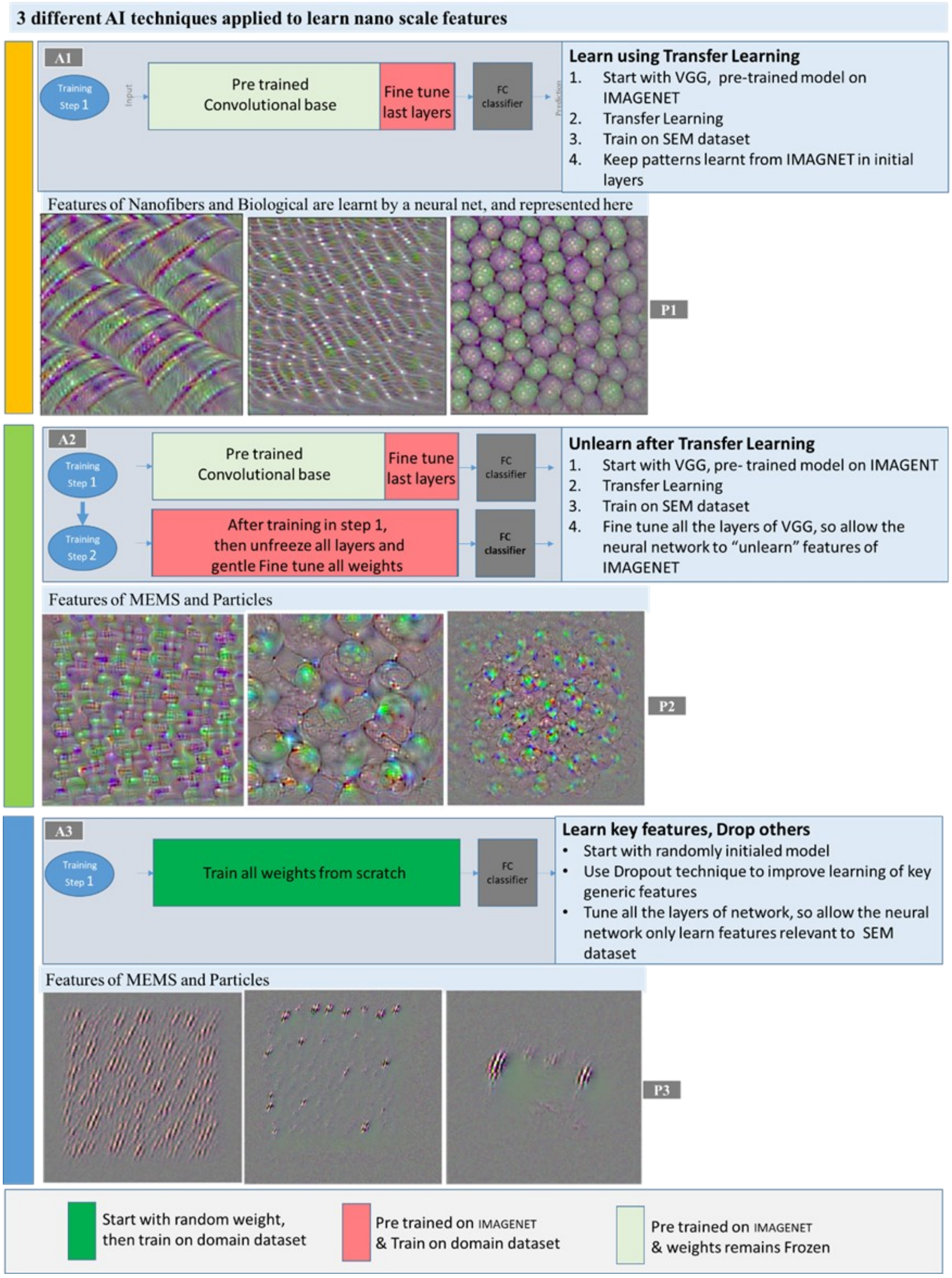
3 different AI techniques applied to learn nano scale features
A1
Training Step 1
Input
Pre trained Convolutional base
Fine tune last layers
FC classifier
Prediction
Learn using Transfer Learning
1. Start with VGG, pre-trained model on IMAGENET
2. Transfer Learning
3. Train on SEM dataset
4. Keep patterns learnt from IMAGNET in initial layers
Features of Nanofibers and Biological are learnt by a neural net, and represented here
P1
A2
Training Step 1
Pre trained Convolutional base
Fine tune last layers
FC classifier
Training Step 2
After training in step 1, then unfreeze all layers and gentle Fine tune all weights
FC classifier
Unlearn after Transfer Learning
1. Start with VGG, pre- trained model on IMAGENT
2. Transfer Learning
3. Train on SEM dataset
4. Fine tune all the layers of VGG, so allow the neural network to "unlearn" features of IMAGENET
Features of MEMS and Particles
P2
A3
Training Step 1
Train all weights from scratch
FC classifier
Learn key features, Drop others
• Start with randomly initialed model
• Use Dropout technique to improve learning of key generic features
• Tune all the layers of network, so allow the neural network only learn features relevant to SEM dataset
Features of MEMS and Particles
P3
Start with random weight, then train on domain dataset
Pre trained on IMAGENET & Train on domain dataset
Pre trained on IMAGENET & weights remains Frozen

**Figure 4:** Methods of training before visualization of architectural patterns. The paper experimented with 3 different methods as explained in this figure. The patterns learnt from IMAGENET were retained by transfer learning method (A2), while patterns of MEMS and NanoParticles were directly learnt in method (A3). The patterns (P2) and (P3) look very different although both the networks were trained on the same dataset of MEMS and Particles.

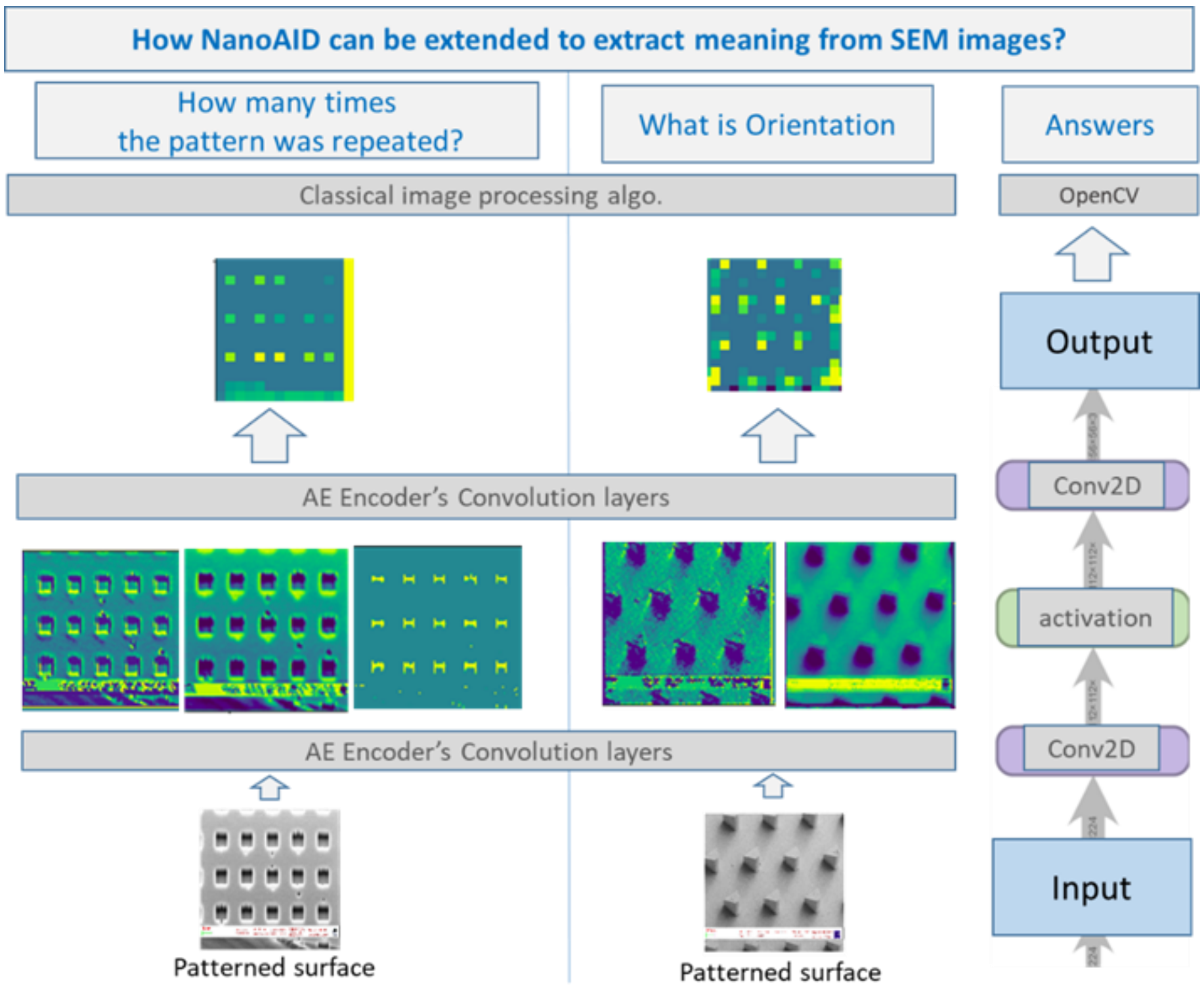

Figure 5: Nanotechnology community can utilize NanoAID to explore the world of nano. Any researcher can tap into the power of AI. For instance, looking an MEMS reveals new structures such as circuits, structures as shown in this AI generated feature maps.

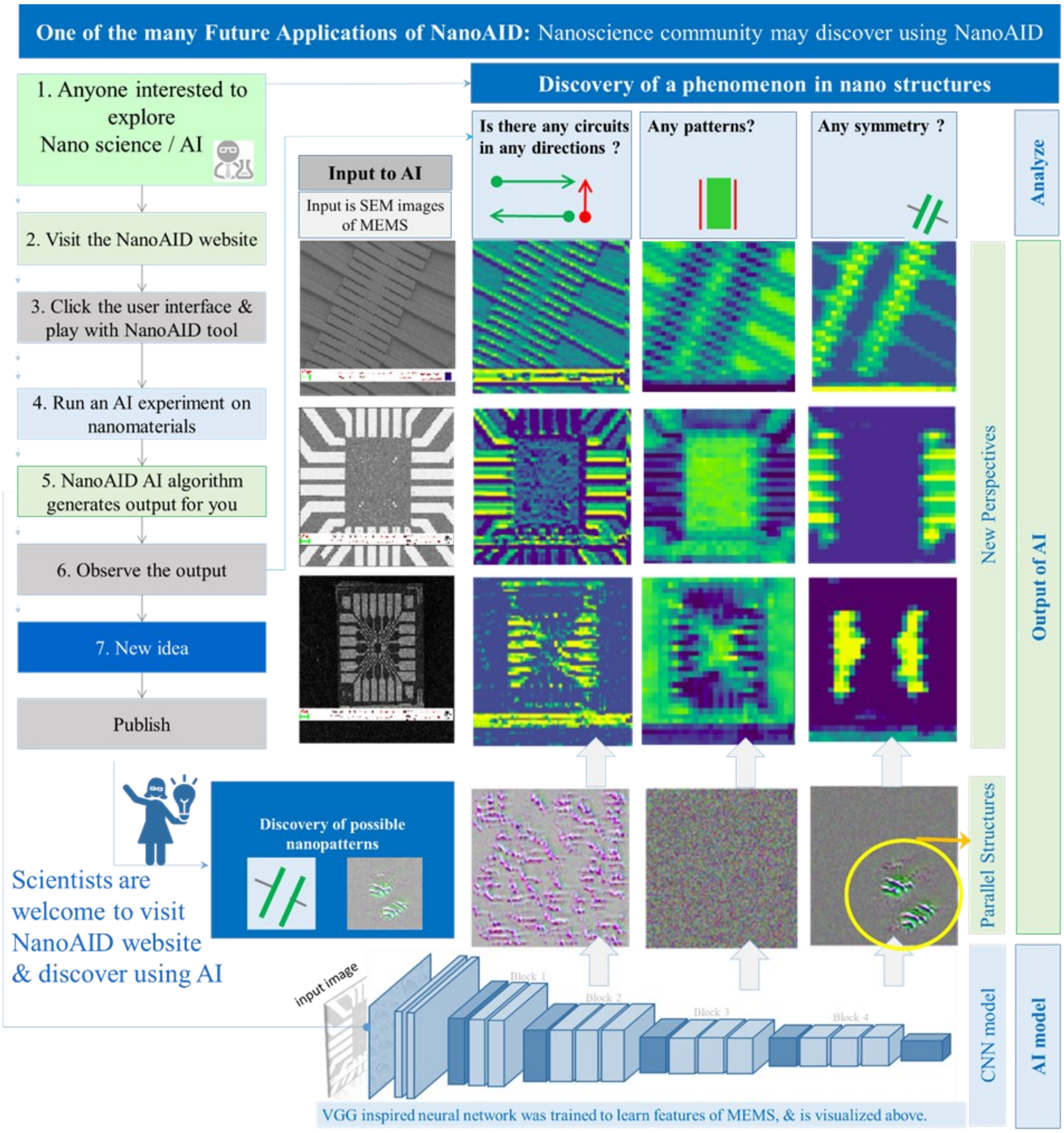

Figure 6: Nanotechnology researchers can extend the Open Source NanoAID. For instance, insights such as count of patterns and its orientation can be extracted from SEM images of Pattern Surfaces.

## 3 Conclusion

For the benefit of community, this work contributed an open source AI platform at the intersection of AI & nanotechnology. NanoAID democratized the power of AI in one of the less published areas of neural net based image transformation in the trending field of Deep Learning in microscopy. This contribution is democratizing access to Deep learning visualization of nano scale materials. The paper provided Google colab URLs to support reproducibility of results and enable community in the potential at the intersection of CNN visualization and nano scale microscopy.

The paper demonstrated how to apply Deep Learning visualization techniques to a public dataset of electron microscopy images of nanomaterials. Novel concepts like "AI lens" was proposed and demonstrated in this work, thus enabling progress in the field of Deep Learning in Microscopy. This paper demonstrated the power of Deep Learning to re-represent a nanomaterial by its features. It was demonstrated that a nanomaterial SEM image could be transformed into its feature representation by computing through a CNN filters of an AutoEncoder. Further, the paper demonstrated experimentally that it is possible to represent a nanomaterial at various levels of abstractions. The "AI lens" also highlighted never seen before patterns.

Extracting nanoscale architecture structural patterns by manual effort from on thousands of images is a challenging science, and this paper also demonstrated a novel approach to do the same using the power of Machine Learning techniques. Three different training techniques were experimented for visualizing the patterns learnt by the model. There is future potential research direction in exploring nanoscience using the power of advances in visualization in representation learning such as the ICLR 2021 research by (Borowski et al., 2021).

The future of AI infused electron microscopy could be in areas such as intelligent visualization to automatically highlight regions and patterns of interest. The paper opened doors to gain new perspectives on nanoscale features by visualizing CNN activation maps of Convolutional AutoEncoders (CAE). The paper demonstrated how Covnet filters in CAE can transform electron microscopy images of nanomaterials, thus helping uncover nanoscale morphologies of different nanomaterials. The visual results generated by this paper show that essentials features can be distilled by an AutoEncoder that is trained to optimize for reconstruction loss by adjusting the dimensions of latent space. Experimenting with the dimensions of latent space can control hierarchical information distillation of various Convolutional layers in an AutoEncoder, thus enabling multiple abstract visual re-representation of the input electron microscopy image.

In short, the potential of AI based visualization in nanotechnology is unlocked in this work and open sourced. The supplementary website of the paper and the source code can be accessed at https://sites.google.com/view/aifornanotechnology .